# NVIDIA AI Aerial: AI-Native Wireless Communications

Kobi Cohen-Arazi, Michael Roe, Zhen Hu, Rohan Chavan, Anna Ptasznik, Joanna Lin, Joao Morais, Joseph Boccuzzi, *Member*, IEEE and Tommaso Balercia

*Abstract*—6G brings a paradigm shift towards AI-native wireless systems, necessitating the seamless integration of digital signal processing (DSP) and machine learning (ML) within the software stacks of cellular networks. This transformation brings the life cycle of modern networks closer to AI systems, where models and algorithms are iteratively trained, simulated, and deployed across adjacent environments.

In this work, we propose a robust framework that compiles Python-based algorithms into GPU-runnable blobs. The result is a unified approach that ensures efficiency, flexibility, and the highest possible performance on NVIDIA GPUs.

As an example of the capabilities of the framework, we demonstrate the efficacy of performing the channel estimation function in the PUSCH receiver through a convolutional neural network (CNN) trained in Python. This is done in a digital twin first, and subsequently in a real-time testbed.

Our proposed methodology, realized in the NVIDIA AI Aerial platform, lays the foundation for scalable integration of AI/ML models into next-generation cellular systems, and is essential for realizing the vision of natively intelligent 6G networks.

# I. INTRODUCTION

With the introduction of 6G, wireless systems will become AI-native frameworks where the digital signal processing and machine learning concepts will be brought to natively coexist. [1] This will bring the development cycle of the cellular network stacks much closer to the software life cycle management typically found in AI system, where components are trained, simulated and deployed in three adjacent environments. A few differences, however, will remain. Control and data planes of cellular systems are expected to remain of similar complexity, despite shrinking time budgets.

The 6G developer will thus be asked to produce complex code, capable of fulfilling very aggressive timelines, while using high-level languages, e.g., Python, which are more convenient while developing AI concepts. This can easily become a complex task, where AI concepts need to be manually integrated. Herein, we propose instead a systematic and programming model, where the final code can be combined with the composition of computational graphs, which run efficiently on the GPU.

The contribution is organized as follows. In Section II, we explain the emerging 3-computer solution and outline the proposed mechanisms to efficiently develop code in such a framework. In Section III, we will cover how the AI Aerial platform implements the 3-computer framework and introduces the system with which we will then demonstrate the proposed framework. Section IV will provide simulated and measured performance results. Section V will deliver the conclusions of this contribution and outline the next steps.

# II. 3-computer framework

As briefly explained before, the 3-computer framework identifies three steps in the development cycle of a given system: design or training, simulation and deployment. Accordingly, it provides three different computers on which each step can be performed.

Figure 1 shows the 3-computer framework. We begin with the design and training phase. Here the AI/ML model is created and trained on a variety of datasets to achieve the desired Key Performance Indicators (KPI).  Once the model shows acceptable values, it is transferred to the simulation environment for validation. For 6G, this typically means a digital twin. [2] If additional tuning is necessary, the model comes back to the training phase. Differently, it is then transferred to a lab deployment using the same Hardware (HW) found in the field for further characterization. If this stage is also successful, the model is deployed in the field, and the Digital Twin can then be used to replay field traces for debugging. Similarly, such traces can be brought to the training phase as well and integrated into the model tuning process to further refine the design. This creates a virtuous circle where the model is continuously validated, deployed and tuned.

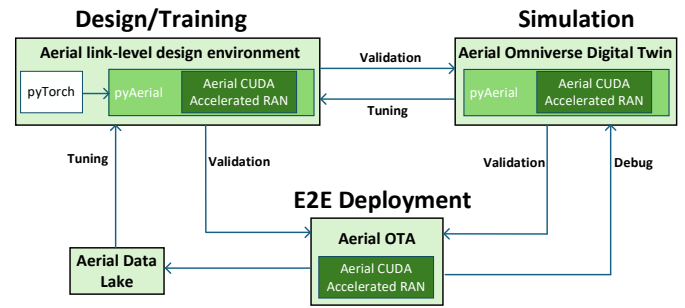

**Fig. 1.** 3-computer development cycle.

## A. Development cycle

The rapid evolution of artificial intelligence and machine learning has driven a demand for development frameworks and methodologies that balance ease of use with high-performance execution. Our approach addresses this by providing a seamless transition from high-level, developer-friendly Python Domain-Specific Languages (DSLs) to highly optimized, low-latency native computational graphs running directly on NVIDIA GPUs. This paradigm significantly accelerates the convergence of neural network (NN) and digital signal processing (DSP) algorithms on NVIDIA GPUs.

### High-Level Development with Python DSLs

The core of our development resides in Python, leveraging its widespread convenience and popularity among academia, researchers, and students. Python's versatility and high-level abstraction capabilities allow developers to express complex ideas quickly, without being burdened by the intricacies of low-level code-to-machine translation. Libraries such as JAX and PyTorch, which serve as powerful Python DSLs, further enhance this usability, making the development of sophisticated models more accessible and efficient. In stark contrast, native languages like C++ are notoriously verbose, harder to write and debug, leading to a more challenging development experience. CUDA, while built on C++, introduces a separate parallel programming model with its own specific concepts (kernels, thread blocks, shared memory, and synchronization), requiring developers to master a new set of paradigms and tools, thus adding a significant layer of complexity to the development process. By enabling development within Python DSLs, we empower users to bypass these complexities, fostering a more expressive and faster development process.

### Transitioning to a High-Performance Native Stack

Our technology bridges the gap between Python's development agility and the imperative for real-time, high-performance execution. We provide users with a robust process to compile their Python-based DSL algorithms into GPU-runnable blobs. A blob is a TensorRT compiled engine binary that is greatly optimized to run on NVIDIA GPU hardware. The binary includes information about the compiled network to be running on the GPU. [3] These blobs are essentially transformed into highly optimized CUDA kernels that can be loaded and executed directly on NVIDIA devices. This transformation facilitates a swift transition from a Python development environment to a real-time, high-performance, low-latency native stack running compiled C++. This transition offers substantial benefits, including the introduction of a quick feedback loop for developers. They can refine their models within their familiar Python DSL environment, compile them into these GPU-runnable blobs, and immediately re-run and re-test them on native, compiled, real-world, real-time systems, potentially over-the-air. This iterative process of measuring performance, refining models, compiling, and re-testing represents a significant advancement in the overall development workflow, creating a virtuous cycle, and drastically shortening the time from NN design to deployment.

### Orchestration of Hybrid Computational Graphs

In a typical development scenario, users working in the Python environment will create multiple NN blobs, not just a single one. Our framework provides a sophisticated mechanism to emit metadata about the entire computational graph, along with crucial information that can be consumed by C++ runtime factories. These factories are responsible for loading the compiled blobs and correctly placing them within the overall larger computational graph.

This larger graph is a hybrid construct, combining hand-crafted, high-performance CUDA compiled C++ files with the optimized TensorRT blobs. All these components work cohesively to form a single, unified computational graph that executes the entire digital signal processing pipeline. This hybrid approach allows us to leverage the strengths of both worlds: the fine-grained control and ultimate performance of custom CUDA C++ kernels for critical DSP stages, the rapid development offered by Python, and the optimized inference capabilities of TensorRT for all types of components.

## B. AI Aerial platform

Underlying the 3-computer solution (provided in figure 1) is the AI Aerial platform, which consists of a design/training environment, simulation environment, and end-to-end deployment on a real network stack. The design and training environment relies on pyAerial, [4] which is a Python API exposing the real-time Layer 1 and Layer 2 functionalities included in the Aerial CUDA Accelerated Radio Access Network (ACAR). [5] The Aerial Data Lake is a tool capable of capturing the front haul I/Q samples from O-RUs, and Layer-2 messages. It consists of the capture application itself, a database of samples collected and an API to access such a database. [6]

The Aerial Omniverse Digital Twin is NVIDIA's environment for system level simulations at scale with hyper realistic radio environment. [7] The Aerial Omniverse Digital Twin offers the possibility of modeling the radio environment in each virtual environment with the use of a deterministic propagation model based on geometrical optics and its extensions. [8] As such it represents a good testing ground to validate that the neural network trained as described above can perform correctly.

Finally, the deployment environment is the Aerial OTA framework, where the Aerial CUDA accelerated RAN is embedded in a 5G full stack and is called Aerial Research Cloud (ARC). [9] Creating an E2E environment consisting of a 5G Stand Alone (SA) Core Network, gNB and UE devices. Together, such blocks realize the NVDIA AI Aerial platform.

# III. NVIDIA AI Aerial

## A. AI Native Example

As an example, we can consider the transmit and receive pipelines for the data channels in figure 2. One of the most critical signal processing blocks in the uplink is channel estimation. In the gNB, this block is used to demodulate the Physical Uplink Shared Channel (PUSCH). [10]

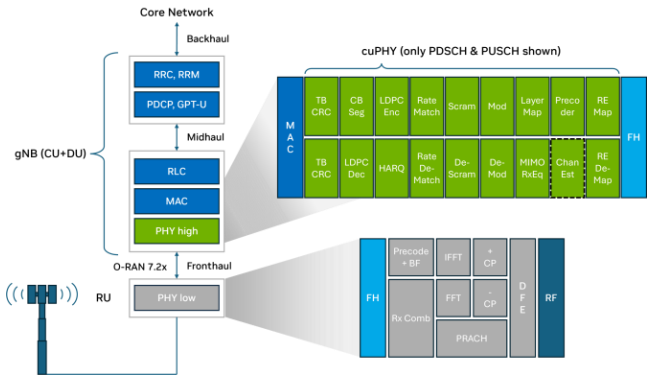

**Fig 2.** Tx/Rx Signal Processing Block Diagram.

Herein, we replace the conventional channel estimation block with a neural network-based approach. We generate a comprehensive training dataset of 28 million channel instances using random realizations of the 3GPP Urban Macro (UMa) channel model in pyAerial, with noisy least squares (LS) channel estimates as input and true channels as targets. [11]

Figure 3 provides an overview of the training workflow. Our training strategy develops a family of models for different operational conditions: separate CNN variants for SNR values from -10 to +40 dB in 5 dB increments, and dedicated models for 1, 4, 16, 64, and 272 PRB configurations. Each CNN architecture features 220k parameters per SNR/PRB tuple with two processing paths: a main channel estimation path and an SNR estimation path. The main path consists of input convolution layers (3×3 kernels), two residual blocks for feature extraction, and output convolution layers that produce denoised channel estimates. The SNR estimation path uses a linear layer to analyze the noisy LS input and estimate signal-to-noise ratio. Input dimensions are structured as (32 conv channels × REs × 2) for real/imaginary components across resource elements (REs) per DMRS symbol.

At inference, the SNR estimation selects the model closest to the actual SNR. For variable PRB allocations, we implement a stitching approach that combines outputs from smaller PRB models to handle uneven resource assignments.

To exemplify the platform's robust support for AI/ML integration, the channel estimation component within the cuPHY software stack has undergone a significant architectural redesign. The PUSCH receiver now features a modular and flexible framework, specifically engineered to allow different signal processing modules, including user-defined channel estimation implementations, to be seamlessly 'plugged in.'

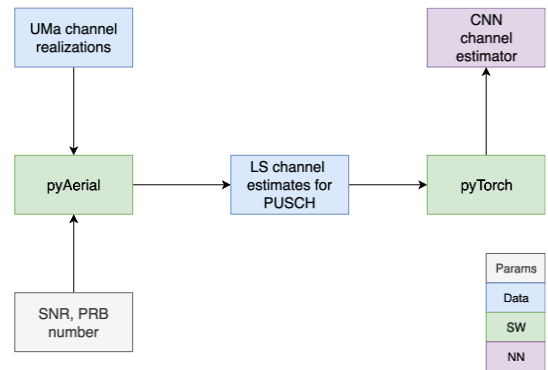

**Fig. 3** Training workflow.

This enhanced architecture adeptly supports both traditional algorithms, such as MMSE, and advanced AI/ML models. Notably, it enables the execution of TensorRT-accelerated neural networks for channel estimation as an integral part of the PUSCH receiver chain, often leveraging fully optimized CUDA graphs for maximum real-time performance. The selection and configuration of the active channel estimation module are managed externally through a straightforward YAML text-based file. This approach not only abstracts the core receiver pipeline from the specifics of the channel estimation technique but also dramatically simplifies the development and deployment lifecycle. AI models can be independently developed, refined (e.g., using Python frameworks), and then deployed as TensorRT engines; the cuPHY system can then load and utilize these new engines without requiring any modifications to the underlying C++ codebase. This accelerates the iteration cycle and facilitates the rapid integration of the latest AI advancements directly into the RAN.

## B. Architectural Options: TorchScript vs. TensorRT Engine

When designing the compilation and deployment pipeline, we evaluated several options for integrating trained neural networks into our C++ runtime. Two primary contenders were TorchScript [12] and NVIDIA TensorRT. While TorchScript offers a way to serialize PyTorch models for C++ inference, it necessitates importing the Libtorch library into our C++ environment. This introduces an external dependency that must be managed, potentially increasing the complexity of deployment and the enforcement of strict timelines. In contrast, we opted for the NVIDIA TensorRT engine as our primary compilation target.

### What is NVIDIA TensorRT?

NVIDIA TensorRT is a Software Development Kit (SDK) for high-performance deep learning inference. It includes a deep learning inference optimizer and runtime that delivers low latency and high throughput for deep learning inference applications. TensorRT takes a trained neural network, which can be from various frameworks like PyTorch or TensorFlow, and optimizes it for deployment. This optimization includes precision calibration (e.g., FP32 to FP16 or INT8), layer fusion, kernel auto tuning, and dynamic tensor memory allocation, resulting in highly efficient execution on NVIDIA GPUs. The output of this optimization process is a TensorRT engine, which is a highly optimized, serialized representation of the neural network graph ready for deployment.

## Advantages of TensorRT Engine Over TorchScript

The decision to utilize the TensorRT engine was driven by several key advantages it offers, particularly in the context of our real-time, low-latency requirements:

- **Reduced Dependencies:** A significant advantage of TensorRT is its tight integration with the CUDA runtime environment. The TensorRT runtime is already baked into the CUDA Toolkit, meaning that deploying a TensorRT engine requires no special imports or additional external dependencies beyond the existing CUDA runtime and driver. This simplifies our C++ compiled runtime significantly, reducing its size and complexity compared to solutions that require external libraries like Libtorch.

## Additional Benefits of Using TensorRT

Beyond the direct comparison with TorchScript, TensorRT offers several other compelling benefits for our high-performance computational graph:

1. **Seamless Graph Integration:** TensorRT engines are designed as computational graphs that can be easily embedded as nodes within a larger computational graph. CUDA provides an intuitive API that allows for capturing the execution stream of a TensorRT engine as a single node, which can then be seamlessly integrated into a broader, more complex computational graph. This modularity is crucial for orchestrating intricate DSP pipelines with strict time requirements.
2. **Optimized Performance:** TensorRT's core purpose is to maximize inference performance on GPUs. Its suite of optimizations, including graph optimizations, kernel selection, and precision calibration, consistently delivers superior throughput and lower latency compared to generic inference solutions or even TorchScript without additional manual optimization.
3. **Flexibility with Pre/Post Kernels:** Our infrastructure provides robust APIs that allow for the dynamic reshaping of tensors by introducing "pre-kernels" and "post-kernels." These are custom CUDA kernels that can be easily injected immediately before and after the TensorRT node within the computational graph. Crucially, these pre/post kernels are highly optimized and very lean, ensuring they do not add any significant overhead to the overall pipeline. This offers immense flexibility and extensibility to the computational pipeline. For instance, if the TensorRT engine expects real-valued inputs but the preceding DSP stage outputs complex numbers, a pre-kernel can unpack these complex numbers into their real and imaginary components. Similarly, a post-kernel can re-pack the output of the TensorRT engine back into complex numbers for subsequent DSP stages. These pre/post kernels are entirely opt-in, providing developers with the choice to use them only when necessary.
4. **Generic Solution:** Our system provides a generic solution that fits any algorithm, not just neural networks. While TensorRT is optimized for NNs, the overall framework defines a clear semantic for input/output (I/O) tensors flowing into and out of the TensorRT blobs. This generic interface allows for the seamless integration of diverse computational elements. Furthermore, the system is generic enough to allow for dynamic switching from a TensorRT node graph to a conventional or traditional non-NN algorithm, providing unparalleled adaptability.

# IV. AI Native Performance

## A. Aerial Omniverse Digital Twin

For this exercise, we have selected a map of Berlin created from open geospatial data from Berlin Partner für Wirtschaft und Technologie GmbH. [13] [14] Figure 4 shows one base station emitting 500,000 rays. The paths shown in the figure represent a subset of the rays that go from the base stations to a set of terminals moving across the map. Each path can present reflections, diffractions and diffusion and is modeled in full polarimetric detail. Antenna array effects, e.g., mutual coupling, are carefully modelled via Active Element Patterns (AEP) loaded at each one of the antenna elements in the arrays at the base and mobile stations moving at speed chosen randomly in the [1.5,2.5] m/s range. This allows us to achieve a high level of accuracy in the characterization of the wireless channel. At the same time, HW-accelerated ray-casting allows us to achieve minimal latencies.

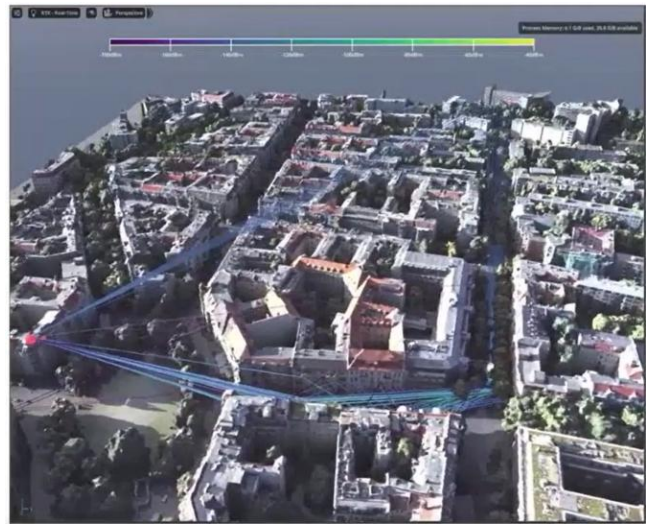

**Fig. 4.** Berlin Germany layout.

Once the channel impulse responses are calculated, they are applied to the wireless channel and, e.g., the PUSCH channel is decoded by the base stations using the Aerial CUDA Accelerated RAN L1. The gNB antenna was placed on the side of a building and a pedestrian UE was placed traveling at the street level. The pedestrian velocity was 5mph.

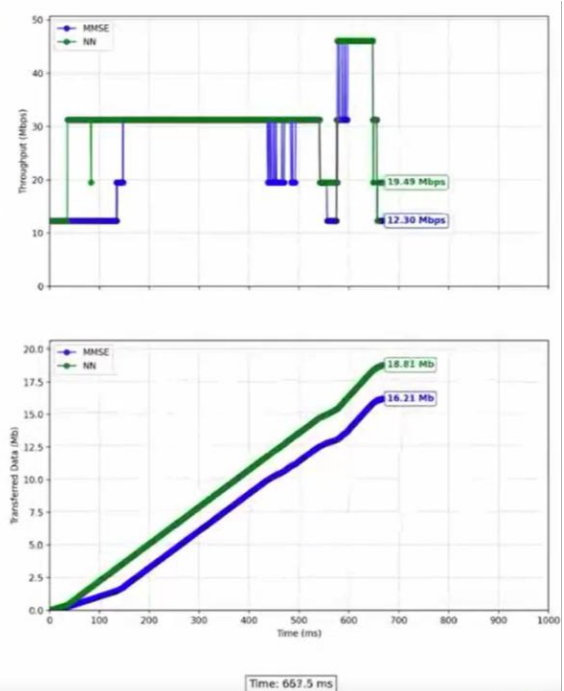

**Fig. 5.** Berlin Germany AODT Simulation results.

Figure 5 provides the AODT simulation results for the Berlin environment. On top, we show the simulation results of the cell instantaneous UL throughput. At the bottom, we show the cumulative UL received data. In this case, we see more than 40% gains in UL throughput when using the CNN model.

## B. Aerial Deployment

Once the Digital Twin system simulations are complete, the next step is to deploy the AI/ML model in an E2E real-time system to evaluate its performance. The E2E system is located on NVIDIA's campus in Santa Clara, CA and is configured as a 5G SA deployment consisting of 2 cells, each supported by a separate gNB.

The 5G SA system consists of a Core Network (CN), gNB with 2x O-CU/O-DU network elements (one for each cell), Front Haul (FH) & Timing distribution network and 2x Radio Units (O-RUs), as shown in Figure 6a. The gNB SW consists of ACAR for Layer 1 and partner stack providers for Layer 2 and above, as well as the CN. Each cell utilized the Multi Instance GPU (MIG) capability.

The proposed ACAR solution inspires the use of a Graphics Processing Unit (GPU), Central Processing Unit (CPU) and Data Processing Unit (DPU). The gNB HW used the Grace Hopper (GH200) server with Blue Field (BF3) DPUs. The Iperf application was used to send UDP traffic. To provide the same wireless channel model experience for each cell, we deployed an eCPRI based system. The FH & Timing distribution network is represented by the PTP Grand Master and Spectrum SN3750 switch. The two O-RUs and UE devices are emulated using the Keysight RuSIM.[15] The RuSIM emulates the O-RU, the wireless channel model and provides the complete UE stack (shown in Figure 6a). The timing reference to both the gNBs and emulated O-RUs is provided by the Precision Time Protocol (PTP) Grand Master from Viavi.

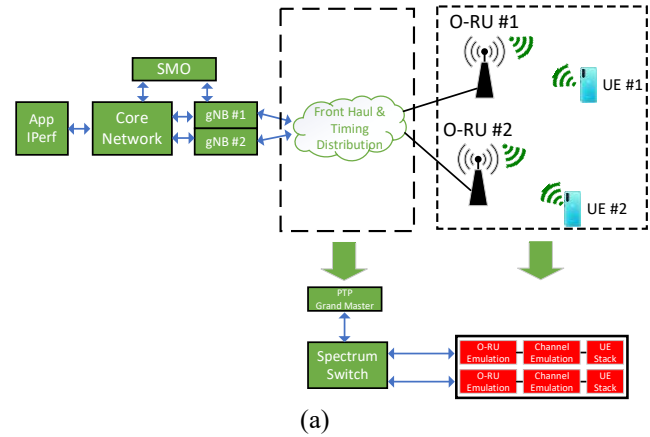

(a)

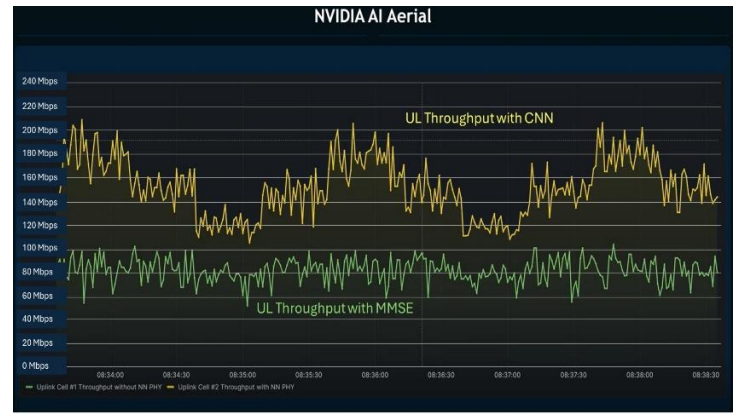

(b)

**Fig. 6.** 5G SA E2E Network Block Diagram (a) and the UL measurement results (b).

The 5G cell configuration consists of a transmission BW = 100MHz using a SCS = 30KHz and capable of using a DL-Heavy TDD slot structure (DDDSUUDDDD) or an UL-Heavy slot structure (DSUUU). Where "D" represents a downlink time slot, "U" represents an uplink time slot and "S" is the special time slot. The traditional MMSE Channel Estimator (CE) used in PUSCH was compared to a CNN model where UL throughput performance is considered. Any improvements in throughput directly increase Spectral Efficiency. Cell #1 has a gNB receiver using the traditional MMSE CE, while Cell #2 has a gNB receiver using the CNN model. Each cell has a single user which has been allocated all 273 PRBs using 1 UL layer. The UL time slot consisted of 10 PUSCH data symbols.

To provide an apples-to-apples comparison, each cell experienced the same wireless channel model, specifically the TDL-C with Delay Spread (DS) = 300nsec. This creates a channel model with the last ray arrives approx. 2.6usec later in time, hence stressing the traditional CE. Each user was traveling at 5mph. The Service Management & Orchestration (SMO) network element is collecting the UL throughput KPIs from both cells and displaying them in a dashboard. The dashboard results are provided in Figure 6b. The SNR was varied to emulate user mobility towards the cell center (when SNR was increasing) and away from the cell center (when SNR was decreasing). The top time series plot shows the UL user throughput connected to gNB #2, while the bottom time series shows the UL user throughput connected to gNB #1. We can see more than 40% gain in UL throughput when using the CNN model in place of the traditional MMSE CE in a TDL-C with DS = 300nsec.

## V. CONCLUSIONS

This generic framework significantly shortens the development time. It achieves this by establishing well-defined contract APIs between the various modules responsible for loading and orchestrating the TensorRT engines. We have introduced dedicated C++ runtime sub-systems as an integral part of this solution, specifically designed to orchestrate these TensorRT engines as nodes within the larger computational graph running on NVIDIA GPUs. This holistic approach ensures efficiency, flexibility, and high performance from development to deployment.

We have provided a new Life Cycle Management (LCM) process to enable the proliferation of AI/ML models into the Telecom network. The LCM consists of the design & training phase, followed by Digital Twin simulation, and then real-world deployment. We have used the traditional channel estimation signal processing algorithm to walk through this framework. By replacing the traditional channel estimation algorithm with a CNN we have shown over 40% improvement in UL throughput in both the Digital Twin as well as in the real-world lab deployment. We believe this is the future of AI-native wireless communications and expect this to be standard for 6G.

## References

[1] AI-RAN Alliance, http://www.ai-ran.org.
[2] X. Lin, L. Kundu, C. Dick, E. Obiodu, T. Mostak and M. Flaxman, "*6G Digital Twin Networks: From Theory to Practice*," IEEE Communications Magazine, vol. 61, no. 11, pp. 72-78, November 2023.
[3] TensorRT SDK | NVIDIA Developer.
[4] pyAerial, pyAerial — Aerial CUDA-Accelerated RAN.
[5] NVIDIA 5G, Aerial CUDA-Accelerated RAN Solution.
[6] Data Lake, Aerial Data Lake documentation.
[7] AODT, Aerial Omniverse Digital Twin documentation.
[8] R. G. Kouyoumjian and P.H. Pathak, "*A uniform geometrical theory of diffraction for an edge in a perfectly conducting surface*," Proceedings of the IEEE, Vol. 62, Issue 11, pp. 1448-1461, November 1974.
[9] Aerial Research Cloud (ARC) 6G, 6G Research Cloud Platform to Advance Wireless Communications With AI.
[10] T. O-Shea and J. Hoydis, "*An Introduction to Deep Learning for the Physical Layer,*" IEEE Transactions on Cognitive Communications and Networking, vol. 3, issue 4, December 2017.
[11] Sionna — Sionna 1.1.0 documentation.
[12] https://docs.pytorch.org/docs/stable/jit.html.
[13] BERLIN PARTNER FUR WIRTSCHAFT UND TECHNOLOGIE GMBH | Enterprise Europe Network.
[14] City of Berlin map, https://download-berlin3d.virtualcitymap.de/.
[15] J. Boccuzzi, R. Chavan, R. Lekhwani, S. Lin, M. Hoang, N. Hedberg, Q. Wang, S. Samala, P. Marini, S. Sarpotdar and M. Adamczyk, "*GPU Accelerated High Capacity, AI-Ready 5G/6G Reference Design and Verification Methodology*," IEEE Communications Magazine, doi: 10.1109/MWC.004.2500023, pp. 1-10, June 2025.

## Biographies

Kobi Cohen-Arazi received a B.Sc. degree in Computer Science from the Academic College of Tel Aviv-Yaffo. He is a Principal Software Engineer with NVIDIA Corporation, focusing on the design and development of next-generation software stacks for 5G/6G Aerial systems.

Michael Roe is a principal engineer at NVIDIA, where he focuses on developing GPU-accelerated software for wireless systems. He holds a Ph.D. in Electrical Engineering from the University of California, Berkeley.

Zhen Hu received the Ph.D. degree in Electrical Engineering from Tennessee Technological University. He is a Senior Software Engineer with NVIDIA focusing on 5G/6G signal processing and CUDA implementation/acceleration.

Rohan Chavan received a Master's degree in Communication Engineering and Media Technology from the University of Stuttgart, Germany. He is a Senior System Software Engineer with NVIDIA corporation focusing on End-to-End features development and integration activities.

Anna Ptasznik is a Senior Software Engineer at NVIDIA focusing on geographic information systems and scene modelling for wireless simulations. She holds an M.S. in Electrical & Computer Engineering from the University of Washington.

Joanna Lin received the M.S. degree in Civil and Environmental Engineering from Stanford University. She is a Senior Software Engineer with NVIDIA Corporation, Santa Clara, CA, responsible for the development of the Aerial Omniverse Digital Twin, with a focus on user interface architecture.

Joao Morais received a Ph.D. in Electrical and Computer Engineering from Arizona State University. He is a Wireless Software Engineer at NVIDIA, focusing on frameworks for AI prototyping in CUDA-accelerated 5G/6G systems.

Joseph Boccuzzi received the Ph.D. degree in Electrical Engineering from New York University, Polytechnic school of engineering, N.Y. He is with NVIDIA Corporation, Santa

Clara, CA, responsible for End-to-End System Architecture for AI Aerial 5G and 6G Wireless Platforms.

Tommaso Balercia is a principal engineer and the lead architect of the Aerial Omniverse Digital Twin at NVIDIA. He holds an M.Sc (Hons.) degree in electrical engineering from Universita' Politecnica delle Marche (Italy) and a Ph.D degree from Technische Universität Braunschweig (Germany). His areas of interest cover computational electromagnetics, digital signal processing, and simulation technology.